\documentclass[conference]{IEEEtran}
\usepackage[hidelinks]{hyperref}
\IEEEoverridecommandlockouts
\usepackage{cite}
\usepackage{amsmath,amssymb,amsfonts}
\usepackage{algorithmic}
\usepackage{graphicx}
\usepackage{textcomp}
\usepackage{xcolor}
\def\BibTeX{{\rm B\kern-.05em{\sc i\kern-.025em b}\kern-.08em
    T\kern-.1667em\lower.7ex\hbox{E}\kern-.125emX}}
\usepackage{booktabs}   
\usepackage{multirow}   
\usepackage{graphicx}   
\usepackage{array}
\usepackage{algorithm}
\usepackage{algorithmic}
\usepackage{amsmath}
\usepackage{graphicx}
\usepackage{placeins}
\begin{document}
\title{RefSR-Adv: Adversarial Attack on Reference-based Image Super-Resolution Models}
\author{\IEEEauthorblockN{Jiazhu Dai\IEEEauthorrefmark{1} and Huihui Jiang}
\IEEEauthorblockA{School of Computer Engineering and Science, Shanghai University, Shanghai 200444, China\\
\IEEEauthorrefmark{1}Correspondence: daijz@shu.edu.cn}}
\maketitle
\begin{abstract}
Single Image Super-Resolution (SISR) aims to recover high-resolution images from low-resolution inputs. Unlike SISR, Reference-based Super-Resolution (RefSR) leverages an additional high-resolution reference image to facilitate the recovery of high-frequency textures. However, existing research mainly focuses on backdoor attacks targeting RefSR, while the vulnerability of the adversarial attacks targeting RefSR has not been fully explored. To fill this research gap, we propose RefSR-Adv, an adversarial attack that degrades SR outputs by perturbing only the reference image. By maximizing the difference between adversarial and clean outputs, RefSR-Adv induces significant performance degradation and generates severe artifacts across CNN, Transformer, and Mamba architectures on the CUFED5, WR-SR, and DRefSR datasets. Importantly, experiments confirm a positive correlation between the similarity of the low-resolution input and the reference image and attack effectiveness, revealing that the model’s over-reliance on reference features is a key security flaw. This study reveals a security vulnerability in RefSR systems, aiming to urge researchers to pay attention to the robustness of RefSR.
\end{abstract}
\begin{IEEEkeywords}
Reference-based Super-resolution, Adversarial Attack
\end{IEEEkeywords}
\section{Introduction}
\label{sec:intro}
Single Image Super-Resolution (SISR) has evolved through various architectures to recover high-resolution details from low-resolution (LR) inputs\cite{dong2015image,liang2021swinir,li2022srdiff, wang2024exploiting}. 
However, due to the lack of sufficient information in low-resolution inputs, SISR inevitably synthesizes unrealistic artifacts or texture hallucinations. To overcome these limitations, Reference-based Super-Resolution (RefSR) has emerged by introducing an high-resolution reference (Ref) image as external high-frequency texture library\cite{zhang2019image, yang2020learning, lu2021masa,jiang2021robust,cao2022reference, zhou2025multi}. 
By leveraging feature matching and fusion, RefSR transfers similar textures from the reference image to achieve superior restoration. Despite it has demonstrated immense potential in security-sensitive domains such as satellite remote sensing \cite{wang2024reference}, medical imaging \cite{kim2025improving}, and intelligent surveillance, the security vulnerabilities of these dual-input systems remain largely unexplored.

Current security research on super-resolution primarily focuses on two dimensions: (i) adversarial attacks on SISR \cite{yin2018deep, choi2019evaluating, reid2025deploying, huang2025scale} by perturbing low-resolution inputs, and (ii) backdoor attacks on RefSR \cite{yang2025badrefsr}, which assume the attacker can contaminate training data. Unlike the single-input architecture of SISR, RefSR possesses a unique dual-input structure (LR and Ref). This architectural characteristic reveals a previously overlooked attack surface: \textbf{Could an attacker exploit the RefSR model's dependence on a reference image to inject subtle perturbations into the reference image to degrade the output?}

In this paper, we systematically expose an inherent security vulnerability in RefSR and propose a novel adversarial attack named \textbf{RefSR-Adv}. Unlike traditional adversarial attack that must tamper with the LR input, RefSR-Adv achieves indirect manipulation by perturbing only the reference image. This strategy offers two core advantages:
\begin{itemize}
    \item \textbf{Integrity of LR Input:} RefSR-Adv maintains the bit-wise integrity of the LR input. In systems where strict integrity audits (e.g., hash verification \cite{ding2023transformer} or digital signatures \cite{korus2017digital}) are deployed on the LR input, traditional attacks fail due to verification errors. RefSR-Adv perfectly bypasses such defenses by ensuring the LR input remains untouched.
    \item \textbf{Enhanced Stealthiness:} In practical workflows, reference images serve as auxiliary inputs and are rarely presented to end-users. Since human scrutiny typically focuses on the final super-resolved result, pixel-level changes in the Ref image are naturally camouflaged and extremely difficult to detect.
\end{itemize}
The primary contributions of this work are summarized as follows:
\begin{enumerate}
    \item We propose RefSR-Adv, revealing the security vulnerability of ``auxiliary surface attacks'' in RefSR systems. To the best of our knowledge, this work represents the first adversarial attack specifically targeting the reference image.
    \item We conduct extensive experiments across four popular RefSR models (CNN, Transformer, and Mamba). Results confirm that this security flaw is universal across different architectures, indicating a general lack of security verification for reference images.
    \item We uncover a positive correlation between the LR-Ref similarity and the performance of the attack, revealing that the excessive reliance on external reference features constitutes a security vulnerability in the RefSR architecture.
\end{enumerate}
\section{Related Work}
\subsection{Image Super-Resolution}
Image Super-Resolution (SR) aims to recover high-resolution (HR) details from low-resolution inputs. Depending on the input sources and prior information utilized, SR can be broadly categorized into SISR and RefSR.

SISR relies on implicit priors learned within the model to reconstruct images from a single LR input. Over the past decade, SISR has evolved from CNNs and Transformers to recent State Space Models (SSMs) and Diffusion Models \cite{dong2015image,liang2021swinir,li2022srdiff, wang2024exploiting}. However, since the information contained in the LR input is inherently limited, SISR models often struggle to reconstruct fine details, leading to unrealistic artifacts or texture hallucinations in the output.

To overcome the inherent information limitations of LR inputs, RefSR incorporates an external high-resolution reference image to migrate high-frequency textures. Through feature matching and adaptive fusion mechanisms, RefSR migrates and transfers similar textures from the Ref image to the reconstructed output, achieving superior detail recovery. The evolution of RefSR has primarily focused on alignment challenges,  progressing from early patch matching \cite{zhang2019image} to Transformer-based mechanisms \cite{yang2020learning, cao2022reference} for enhanced robustness against disparity. Recently, \cite{zhou2025multi} integrated the Mamba architecture for efficient long-range dependency modeling. While recent works like RefDiff \cite{dong2024building} explore dual-input diffusion models, their stochastic denoising mechanisms fundamentally differ from the deterministic feature mapping used in CNN, Transformer, and SSM architectures, this study specifically focuses on the security vulnerabilities in these deterministic architectures.

\subsection{Security Threats in Super-Resolution}
Security research in image super-resolution primarily investigates two distinct threat categories: Adversarial Attacks and Backdoor Attacks.

Adversarial attacks aim to induce catastrophic performance degradation by introducing subtle, intentionally designed perturbations into the input data during the inference phase. Early pioneering work \cite{choi2019evaluating} systematically evaluated the vulnerability of various SISR architectures, while \cite{yin2018deep} revealed that adversarial attacks on SISR can serve as ``upstream interference'' to mislead downstream tasks. Subsequently, for complex scenarios, SIAGT \cite{huang2025scale} achieved scale-invariant attacks, and \cite{reid2025deploying} explored the deployment challenges of adversarial samples in edge device inference streams. However, current adversarial research in super-resolution primarily concentrates on compromising single-input SISR models by perturbing the low-resolution (LR) stream. Due to the unique dual-input architecture of RefSR, which integrates both LR and Ref features, the vulnerability of the reference path to adversarial attack remains entirely unexplored. To fill this gap, RefSR-Adv introduces a adversarial attack that targets the previously overlooked "auxiliary surface". By injecting subtle perturbations into the reference image, our framework successfully induces catastrophic output degradation.

Backdoor attacks involve embedding hidden malicious behaviors into a model by injecting triggers into the training dataset, a process known as "data poisoning". Recent research, BadRefSR\cite{yang2025badrefsr}, has explored this threat in RefSR systems by adding triggers to reference images during the training phase. While these studies highlight significant risks, they assume the attacker has the capability to contaminate training data, which may not be feasible in many real-world scenarios. Unlike backdoor-based "data poisoning," RefSR-Adv operates as an adversarial threat during the deployment or inference process, requiring no access to the training phase. While backdoor threats have been investigated, the adversarial attacks targeting the reference image during the inference process remains unexplored. RefSR-Adv fills this research gap.

\section{METHODOLOGY}
\label{sec:method}
In this section, we first provide a formal definition of RefSR. We then analyze the limitations of existing attacks on SISR and propose our threat model. Finally, we elaborate on the optimization objectives and algorithmic details of the RefSR-Adv attack.
\subsection{Preliminary}
Unlike SISR, which relies on implicit priors within the model for reconstruction, RefSR introduces a high-resolution reference image $I_{Ref}$ as an external high-frequency texture library. Formally, given a low-resolution input $I_{LR} \in \mathbb{R}^{H \times W \times C}$ containing the primary structure and a reference image $I_{Ref} \in \mathbb{R}^{H_{ref} \times W_{ref} \times C}$ providing detail priors, the RefSR model $\mathcal{M}$ aims to reconstruct a high-resolution image $I_{SR} \in \mathbb{R}^{sH \times sW \times C}$ ($s$ is the upsampling factor):
\begin{equation}
I_{SR} = \mathcal{M}(I_{LR}, I_{Ref}; \theta),
\end{equation}
where the parameters $\theta$ are typically optimized via one of two mainstream strategies:
\begin{itemize}
    \item \textbf{Reconstruction-only ($L_{rec}$):} This strategy focuses on ensuring pixel-level signal fidelity. The reconstruction loss is typically formulated using the $L_1$-norm to measure the absolute discrepancy between the super-resolved output and the ground-truth $I_{GT}$ image:
    \begin{equation}
        L_{rec} = \frac{1}{N} \sum_{i=1}^{N} \| \mathcal{M}(I_{LR}^i, I_{Ref}^i; \theta) - I_{GT}^i \|_1,
    \end{equation}
    where $N$ is the number of training samples. While optimization under this objective yields high numerical scores in terms of PSNR and SSIM, the individual $L_1$ loss tends to cause over-smoothed results that lack fine-grained textures.

    \item \textbf{Full-loss ($L_{full}$):} To improve perceptual quality and generate more visually favorable details, a composite total loss is employed: $L_{full} = L_{rec} + \lambda_{1} L_{per} + \lambda_{2} L_{adv}$.The hyperparameters $\lambda_{1}$ and $\lambda_{2}$ are used as balancing coefficients to adjust the trade-off between pixel-level signal fidelity and higher-level perceptual realism.
    
    \textit{Perceptual Loss ($L_{per}$):} By utilizing feature maps from a pre-trained VGG model, $L_{per}$ constrains the model in a high-dimensional feature space:
    \begin{equation}
            L_{per} = \frac{1}{N} \sum_{i=1}^{N} \| \phi_{j}(I_{SR}^i) - \phi_{j}(I_{GT}^i) \|_F,
    \end{equation}
    where $\phi_{j}(\cdot)$ denotes the $j$-th layer output of the VGG model and $\|\cdot\|_F$ denotes the Frobenius norm.
    
    \textit{Adversarial Loss ($L_{adv}$):} Typically implemented via Generative Adversarial Networks (GANs), this loss encourages the model to synthesize realistic high-frequency textures by penalizing the distribution gap between generated and real images:
    \begin{equation}
        L_{adv} = -\mathbb{E}_{I_{SR}} [\log(D(I_{SR}))],
    \end{equation}
    where $D$ is the discriminator tasked with distinguishing real ground-truth images from reconstructed ones.
    This strategy significantly enhances the model's ability to migrate and reconstruct intricate textures, but it potentially increases the network's sensitivity and "excessive trust" toward reference features.
\end{itemize}
\subsection{Threat Model and Problem Formulation}
In this study, we investigate the adversarial robustness of RefSR models under a white-box attack setting, which serves as a rigorous evaluation of the model's security boundary.
\subsubsection{Attacker Capability}
Following the standard adversarial settings in super-resolution research\cite{yin2018deep, choi2019evaluating, reid2025deploying, huang2025scale}, we assume the attacker has full knowledge of the target RefSR model $\mathcal{M}$, including its specific architecture, internal parameters $\theta$, and the gradients required for optimization. The attacker’s capability is confined to injecting a subtle, pixel-level adversarial perturbation $\delta$ into the high-resolution reference image $I_{Ref}$, while the primary low-resolution input $I_{LR}$ remains unmodified.
\subsubsection{Problem Formulation}
The objective of RefSR-Adv is to identify an optimal  adversarial perturbation $\delta$ that, when added to the reference image, induces the maximum reconstruction error in the super-resolved output. Let $I_{GT}$ represent the ground-truth high-resolution image. We formulate the attack as a constrained optimization problem aimed at maximizing the loss between the model's output and the ground truth:
\begin{equation}\max_{\delta} \mathcal{L} \Big( \mathcal{M}(I_{LR}, I_{Ref} + \delta), I_{GT} \Big),\end{equation}
subject to the following constraints:\begin{equation}||\delta||_{\infty \le} \epsilon, \quad (I_{Ref} + \delta) \in [0, 1]^{H_{ref} \times W_{ref} \times C},\end{equation}
where $\mathcal{L}(\cdot)$ denotes a loss function (e.g., $L_2$ loss) utilized to quantify the degradation in signal fidelity. The term $\epsilon$ signifies the maximum allowable perturbation budget, ensuring that the adversarial modifications remain imperceptible to human observers.

\subsection{RefSR-Adv Attack}

\begin{figure}[t]
    \centering
    \includegraphics[width=0.9\linewidth]{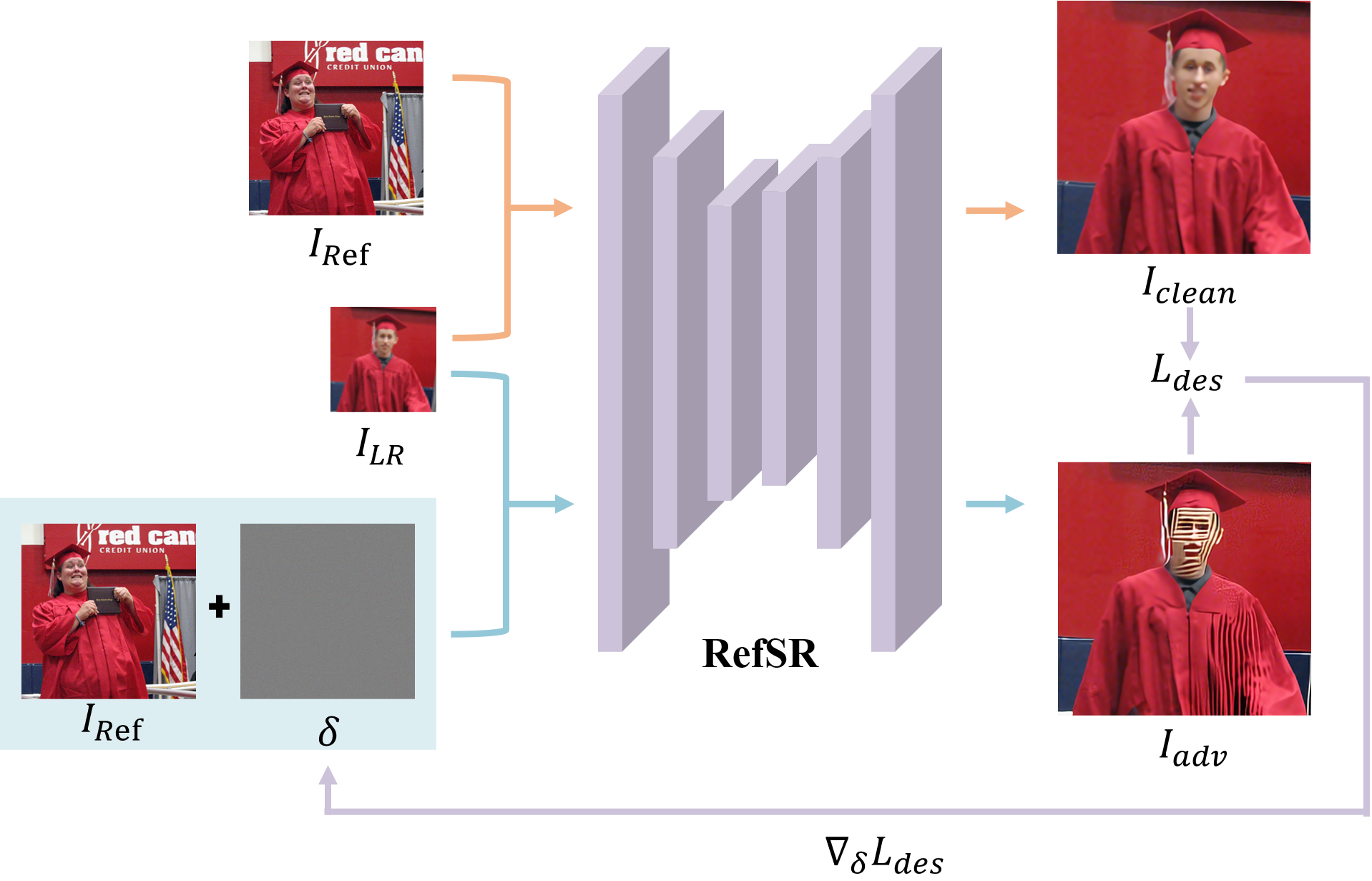}
    \caption{Overview of the RefSR-Adv attack framework. The procedure comprises two primary stages: (1) \textbf{Baseline Generation}, where the clean super-resolution output $I_{clean}$ is synthesized to serve as the pseudo ground-truth anchor; and (2) \textbf{Adversarial Optimization}, where a learnable perturbation $\delta$ is iteratively optimized within the auxiliary reference stream to maximize the output discrepancy, ultimately inducing severe textural artifacts in the final adversarial output $I_{adv}$.}
    \label{fig:framework}
\end{figure}

As shown in Fig. \ref{fig:framework}, RefSR-Adv employs a gradient-based iterative optimization paradigm consisting of three core components:

\subsubsection{Pseudo Ground-Truth Strategy}
In practical inference scenarios, the actual high-resolution ground-truth image $I_{\text{GT}}$ is inherently unavailable to the attacker. To address this, we adopt a \textit{pseudo ground-truth strategy} \cite{yin2018deep, choi2019evaluating}, utilizing the model's own output under benign conditions as the reference baseline. Specifically, we define the clean super-resolution output, generated from the original low-resolution image $I_{LR}$ and the clean reference image $I_{Ref}$, as the baseline:
\begin{equation}
    I_{clean} = \mathcal{M}(I_{LR}, I_{Ref}; \theta).
\end{equation}
By treating $I_{clean}$ as a high-fidelity proxy for the intended reconstruction, we can precisely quantify the degree of adversarial deviation. This strategy ensures the attack's effectiveness in real-world deployment environments where the ground-truth is unknown, providing a stable “intended” baseline for optimization.

\subsubsection{Destruction Loss Formulation}
To induce maximum degradation in signal fidelity, we formulate a \textit{destruction loss} $\mathcal{L}_{des}$ aimed at maximizing the discrepancy between the adversarial output $I_{adv}$ and the clean baseline $I_{clean}$. Let $I_{adv} = \mathcal{M}(I_{LR}, I_{Ref} + \delta; \theta)$ denote the output generated from the perturbed reference image. We utilize the $L_2$ norm to formalize the objective:
\begin{equation}
    \mathcal{L}_{des}(\delta) = \| I_{adv} - I_{clean} \|_2.
    \label{eq:loss_des}
\end{equation}
The choice of the $L_2$ norm is motivated by two key factors. First, maximizing the Euclidean discrepancy effectively disrupts the pixel-level reconstruction consistency inherent in deterministic architectures such as CNN, Transformer and Mamba. Second, since maximizing the Mean Squared Error (MSE) is mathematically equivalent to minimizing the Peak Signal-to-Noise Ratio (PSNR), the $L_2$ norm serves as a robust and natural proxy for inducing catastrophic reconstruction error.

\subsubsection{Optimization via Projected Gradient Descent}
To solve the constrained maximization problem defined by the destruction loss, we employ the Projected Gradient Descent (PGD) algorithm \cite{madry2017towards}. Unlike simpler methods, PGD utilizes \textit{random initialization} to more comprehensively explore the adversarial loss landscape within the perturbation budget $\epsilon$. In each iteration $t$, the learnable perturbation $\delta$ is updated along the direction of the gradient sign:
\begin{equation}
    \delta^{(t+1)} = \Pi_{\epsilon} \left[ \delta^{(t)} + \alpha \cdot \text{sign}\left( \nabla_{\delta} \mathcal{L}_{des}(\delta^{(t)}) \right) \right],
\end{equation}
where $\alpha$ denotes the step size and $\Pi_{\epsilon}(\cdot)$ represents the projection operator ensuring the perturbation remains within the $\ell_\infty$-norm constraint $\|\delta\|_\infty \le \epsilon$ and valid pixel range $[0, 1]$. By exploiting the differentiable nature of modern texture matching and fusion modules, RefSR-Adv backpropagates output discrepancies directly to the reference image pixels to identify the most damaging perturbations. The complete optimization logic is summarized in Algorithm \ref{alg:pgd}.

\begin{algorithm}[ht]
\caption{RefSR-Adv: Reference-based Adversarial Perturbation Optimization}
\label{alg:pgd}
\begin{algorithmic}[1]
\REQUIRE Target RefSR model $\mathcal{M}$ with parameters $\theta$; Clean primary input $I_{LR}$; Clean auxiliary reference image $I_{Ref}$; Perturbation budget $\epsilon$; Step size $\alpha$; Total iterations $T$.
\ENSURE Adversarial reference image $I_{Ref}^{adv}$.

\STATE \textbf{Step 1: Baseline Generation}
\STATE $I_{clean} \leftarrow \mathcal{M}(I_{LR}, I_{Ref}; \theta)$ 

\STATE \textbf{Step 2: Perturbation Initialization}
\STATE $\delta^{(0)} \leftarrow \text{Uniform}(-\epsilon, \epsilon)$ 
\STATE $\delta^{(0)} \leftarrow \text{Clip}(I_{Ref} + \delta^{(0)}, 0, 1) - I_{Ref}$ 

\STATE \textbf{Step 3: Iterative Adversarial Optimization}
\FOR{$t = 0$ \TO $T-1$}
    \STATE $I_{adv} \leftarrow \mathcal{M}(I_{LR}, I_{Ref} + \delta^{(t)}; \theta)$ 
    \STATE $\mathcal{L}_{des} \leftarrow \| I_{adv} - I_{clean} \|_2$ 
    \STATE $G \leftarrow \nabla_{\delta} \mathcal{L}_{des}(\delta^{(t)})$ 
    \STATE $\delta^{(t+1)} \leftarrow \delta^{(t)} + \alpha \cdot \text{sign}(G)$ 
    \STATE $\delta^{(t+1)} \leftarrow \text{Clip}(\delta^{(t+1)}, -\epsilon, \epsilon)$ 
    \STATE $\delta^{(t+1)} \leftarrow \text{Clip}(I_{Ref} + \delta^{(t+1)}, 0, 1) - I_{Ref}$ 
\ENDFOR

\RETURN $I_{Ref}^{adv} = I_{Ref} + \delta^{(T)}$
\end{algorithmic}
\end{algorithm}

\section{EXPERIMENTS}
\label{sec:exp}
In this section, we conduct quantitative and qualitative evaluations to assess the effectiveness and stealthiness of RefSR-Adv. We first describe the experimental setup, followed by a performance analysis across four popular RefSR models to demonstrate the universality of the identified vulnerabilities.
\subsection{Experimental Settings}
\subsubsection{Datasets} We evaluate our method on three standard datasets: 
\begin{itemize}
    \item \textbf{CUFED5\cite{yang2020learning}}, featuring 126 groups with varying reference similarity levels; 
    \item \textbf{WR-SR\cite{jiang2021robust}} , containing web-crawled images with diverse viewpoints and lighting to simulate real-world scenarios; 
    \item \textbf{DRefSR\cite{zhou2025multi}} , focused on diverse texture exploitation across categories like architecture and animals. 
\end{itemize} 
To balance computational efficiency with detail preservation, we adopt a $600 \times 600$ center-cropping strategy for high-resolution datasets (WR-SR and DRefSR).
\subsubsection{Victim Models}
To verify the universality of RefSR-Adv, we select four popular models covering three mainstream paradigms (CNN, Transformer, Mamba) :
\begin{itemize}
\item \textbf{TTSR\cite{yang2020learning}}: A pioneering \textbf{Transformer-based} RefSR model that utilizes ``Hard-Soft Attention" mechanisms to improve the accuracy of texture feature transfer from Ref images.
\item \textbf{MASA-SR\cite{lu2021masa}}: A classic \textbf{CNN-based} representative that employs spatial adaptation modules and coarse-to-fine matching to significantly enhance feature alignment efficiency.
\item \textbf{DATSR\cite{cao2022reference}}: An advanced \textbf{Transformer} architecture that adopts Deformable Attention to achieve robust feature matching and detail recovery, especially under large parallax conditions.
\item \textbf{SSMTF\cite{zhou2025multi}}: The latest \textbf{Mamba-based} model that leverages State Space Models for efficient long-range dependency modeling and multi-scale texture fusion.
\end{itemize}
\subsubsection{Evaluation Metrics}
We utilize standard SR metrics: Peak Signal-to-Noise Ratio (PSNR) and Structural Similarity Index (SSIM).  
\begin{itemize}
    \item PSNR measures the pixel-level reconstruction fidelity based on the Mean Squared Error.
    \item SSIM evaluates the structural similarity by considering luminance, contrast, and texture information.
\end{itemize}
We record three categories of results: (i) SR quality under clean references; (ii) SR quality under adversarial references; (iii) Fidelity of the adversarial reference relative to the clean image to measure \textbf{Stealthiness}.
\subsubsection{Implementation Details} We employ the PGD optimizer with a perturbation budget of $\epsilon=8/255$ and $T=50$ iterations to generate adversarial samples. All experiments are conducted for $4\times$ super-resolution. Notably, since all victim models are re-implemented locally using official source codes, the baseline performances may exhibit discrepancies from the results reported in the original papers.
\subsection{Attack Performance Evaluation}
To evaluate the effectiveness of our framework, we conduct quantitative assessments of RefSR-Adv across four state-of-the-art models on three standard datasets. Table \ref{tab:main_results} illustrates the quantitative impact of the proposed attack. 
\begin{table*}[h]
\centering
\caption{\footnotesize The attack performance (PSNR/SSIM) of RefSR-Adv on various RefSR models across three datasets, where the ‘Clean Output’ and ‘Adversarial Output’ columns represent the SR quality using original references and adversarial references, The ‘Performance Drop’ denotes the degradation, and the ‘Stealthiness’ column represents the PSNR/SSIM between the original reference image and the adversarial reference. The suffix '-rec' denotes models trained with only reconstruction loss.}
\label{tab:main_results}
\setlength{\tabcolsep}{8pt} 
\renewcommand{\arraystretch}{0.8} 
\resizebox{0.9\textwidth}{!}{
\begin{tabular}{l|l|c|c|c|c}
\toprule
\textbf{Dataset} & \textbf{Model} & \textbf{Clean Output} & \textbf{Adversarial Output} & \textbf{Performance Drop } & \textbf{Stealthiness} \\
\midrule
\multirow{8}{*}{\textbf{CUFED5}} 
 & TTSR      & 25.40 / 0.7600 & 21.84 / 0.5599 & 3.56 / 0.2001 & 35.71 / 0.9138 \\
 & TTSR-rec  & 26.99 / 0.8003 & 23.55 / 0.7145 & 3.44 / 0.0858 & 36.28 / 0.9236 \\
 & MASA      & 24.65 / 0.7257 & 19.69 / 0.5811 & 4.96 / 0.1446 & 36.99 / 0.9316 \\
 & MASA-rec  & 27.35 / 0.8140 & 24.55 / 0.7549 & 2.80 / 0.0591 & 37.03 / 0.9349 \\
 & \textbf{DATSR} & \textbf{27.76 / 0.8285} & \textbf{17.12 / 0.4690} & \textbf{10.64 / 0.3595} & \textbf{36.15 / 0.9236} \\
 & DATSR-rec & 28.49 / 0.8510 & 18.35 / 0.5640 & 10.14 / 0.2870 & 36.05 / 0.9232 \\
 & SSMTF     & 28.13 / 0.8383 & 18.88 / 0.5569 & 9.25 / 0.2814 & 35.75 / 0.9193 \\
 & SSMTF-rec & 28.77 / 0.8553 & 19.31 / 0.6290 & 9.46 / 0.2263 & 36.19 / 0.9249 \\
\midrule
\multirow{8}{*}{\textbf{WR-SR}}  
 & TTSR      & 26.38 / 0.7480 & 21.60 / 0.4809 & 4.78 / 0.2671 & 35.80 / 0.9078 \\
 & TTSR-rec  & 27.53 / 0.7803 & 23.55 / 0.6680 & 3.98 / 0.1123 & 36.29 / 0.9163 \\
 & MASA      & 25.33 / 0.7027 & 20.51 / 0.5795 & 4.82 / 0.1232 & 37.30 / 0.9294 \\
 & MASA-rec  & 27.72 / 0.7836 & 25.68 / 0.7594 & 2.04 / 0.0242 & 37.47 / 0.9342 \\
 & \textbf{DATSR} & \textbf{27.39 / 0.7732} & \textbf{16.76 / 0.5353} & \textbf{10.63 / 0.2379} & \textbf{36.63 / 0.9214} \\
 & DATSR-rec & 27.83 / 0.7916 & 18.41 / 0.6122 & 9.42 / 0.1794 & 36.52 / 0.9200 \\
 & SSMTF     & 27.51 / 0.7767 & 19.80 / 0.6292 & 7.71 / 0.1475 & 36.25 / 0.9164 \\
 & SSMTF-rec & 27.89 / 0.7929 & 19.89 / 0.6680 & 8.00 / 0.1249 & 36.71 / 0.9224 \\
\midrule
\multirow{8}{*}{\textbf{DRefSR}} 
 & TTSR      & 28.06 / 0.7886 & 22.94 / 0.5299 & 5.12 / 0.2587 & 35.78 / 0.8982 \\
 & TTSR-rec  & 29.28 / 0.8175 & 25.00 / 0.7207 & 4.28 / 0.0968 & 36.24 / 0.9070 \\
 & MASA      & 27.03 / 0.7500 & 20.81 / 0.6050 & 6.22 / 0.1450 & 37.31 / 0.9237 \\
 & MASA-rec  & 29.47 / 0.8213 & 26.68 / 0.7839 & 2.79 / 0.0374 & 37.39 / 0.9270 \\
 & \textbf{DATSR} & \textbf{29.37 / 0.8161} & \textbf{17.36 / 0.4760} & \textbf{12.01 / 0.3401} & \textbf{36.41 / 0.9093} \\
 & DATSR-rec & 29.95 / 0.8347 & 19.30 / 0.6180 & 10.65 / 0.2167 & 36.35 / 0.9097 \\
 & SSMTF     & 29.54 / 0.8221 & 20.13 / 0.6160 & 9.41 / 0.2061 & 36.11 / 0.9066 \\
 & SSMTF-rec & 30.06 / 0.8380 & 20.24 / 0.6736 & 9.82 / 0.1644 & 36.55 / 0.9127 \\
\bottomrule
\end{tabular}%
}
\end{table*}
As shown in the results, while TTSR and MASA exhibit relative robustness, DATSR and SSMTF suffer severe performance collapses, with PSNR drops often exceeding 7dB. This discrepancy stems from their specific feature-matching strategies. TTSR and MASA downsample reference images to handle scale disparities; mathematically, this acts as a low-pass filter that inadvertently mitigates high-frequency perturbations. Conversely, DATSR and SSMTF interact with features at original resolutions to pursue superior detail recovery. Without the filtering protection, these models fully absorb and amplify adversarial textures, leading to catastrophic degradation. 

Furthermore, a comparative analysis of different training objectives reveals that models optimized with the full-loss function ($L_{full}$) generally exhibit higher vulnerability to RefSR-Adv than those trained with reconstruction-only ($L_{rec}$) objectives, particularly for the TTSR, MASA, and DATSR. While perceptual and adversarial losses ($L_{per}$ and $L_{adv}$) are designed to encourage the synthesis of realistic high-frequency textures, RefSR-Adv strategically exploits this mechanism by misleading the network to misinterpret adversarial noise as valid textural details, thereby inducing severe visual artifacts. Conversely, the inherent tendency of $L_{rec}$-optimized models toward over-smoothed reconstructions provides a natural suppression mechanism against such high-frequency perturbations. However, SSMTF presents a notable exception where the reconstruction-only version suffers a slightly more pronounced performance drop than its full-loss counterpart. This phenomenon is attributed to Mamba’s unique global state evolution, which causes pixel-level perturbations to propagate and accumulate throughout the entire sequence when the model is constrained by strict pixel-level fidelity. In this specific case, the high-level semantic regularization provided by the full-loss objective functions as a robust buffer, effectively mitigating the global amplification of low-level adversarial noise.

Overall, these results demonstrate that RefSR-Adv maintains high stealthiness (PSNR $>$ 35dB) to ensure that adversarial perturbations remain imperceptible. The significant performance degradation reveals a universal security vulnerability across mainstream CNN, Transformer, and Mamba architectures. This fundamental flaw stems from the models' excessive reliance on untrusted reference images, proving that the auxiliary reference stream constitutes a critical and vulnerable attack surface. 
\subsection{Qualitative Analysis}
\begin{figure}[h]
\centering
\includegraphics[width=\linewidth]{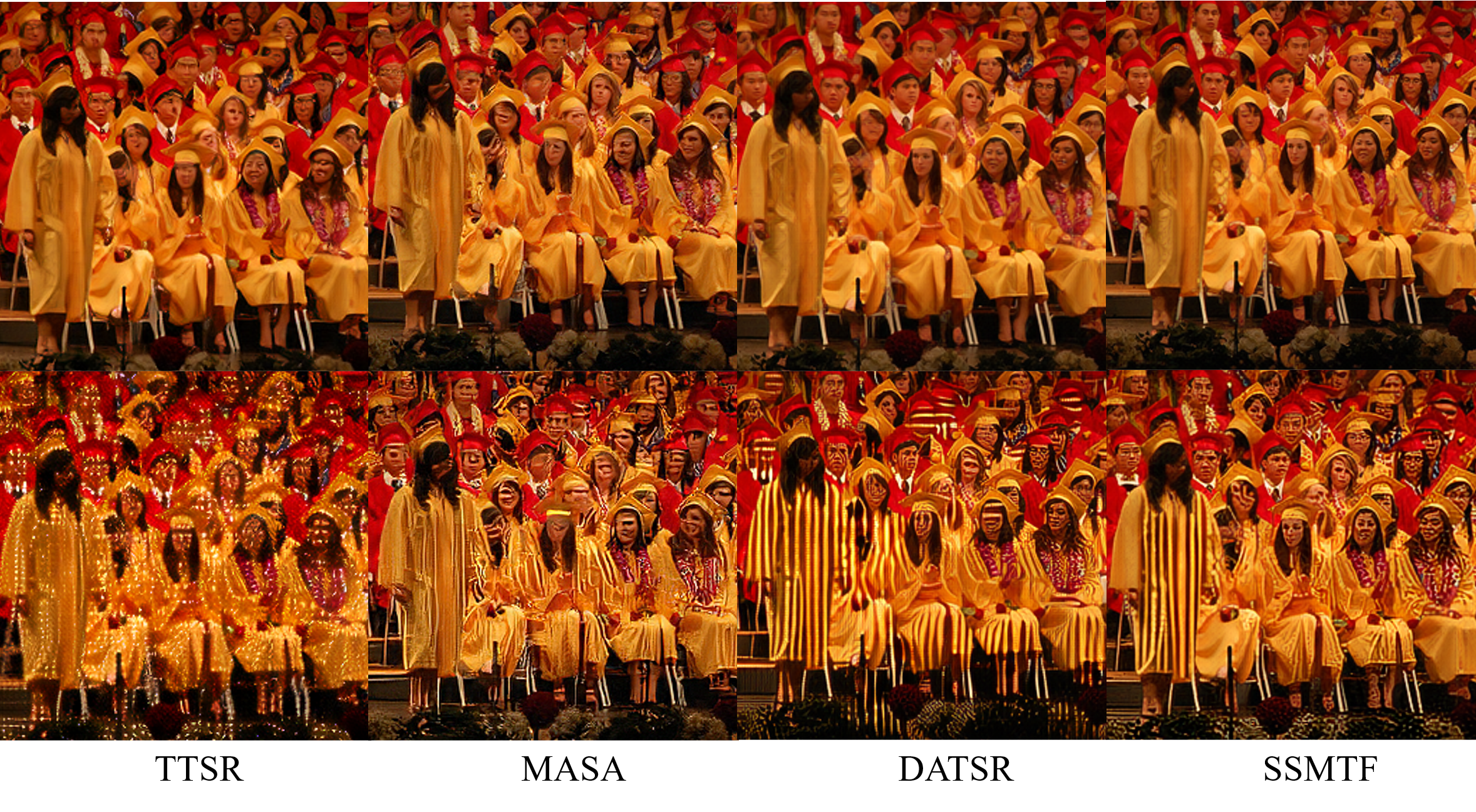}
\caption{Visual results on CUFED5 dataset. }
\label{fig:vis_cufed}
\end{figure}

\begin{figure}[h]
\centering
\includegraphics[width=\linewidth]{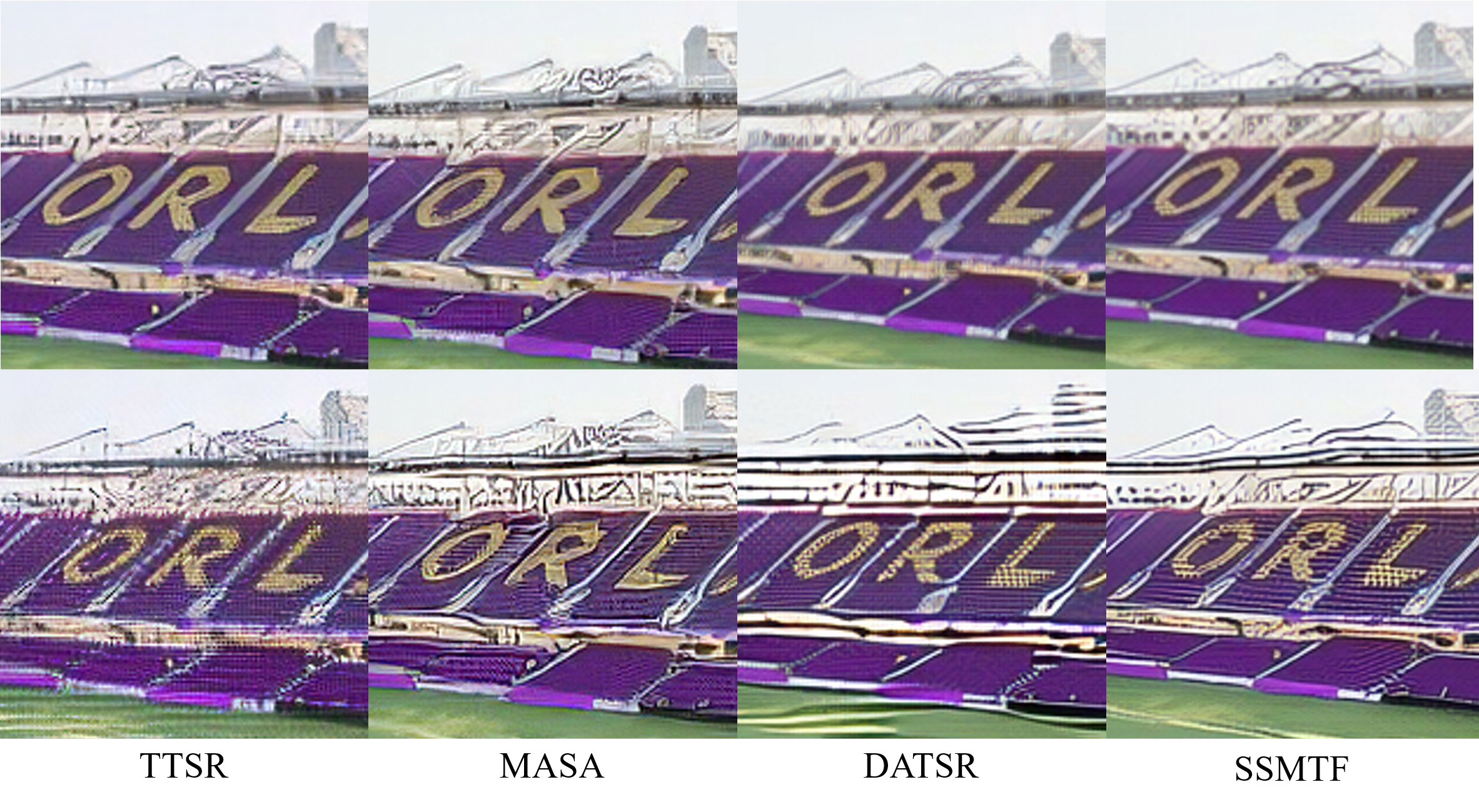}
\caption{Visual results on WR-SR dataset.}
\label{fig:vis_wr}
\end{figure}

\begin{figure}[h]
\centering
\includegraphics[width=\linewidth]{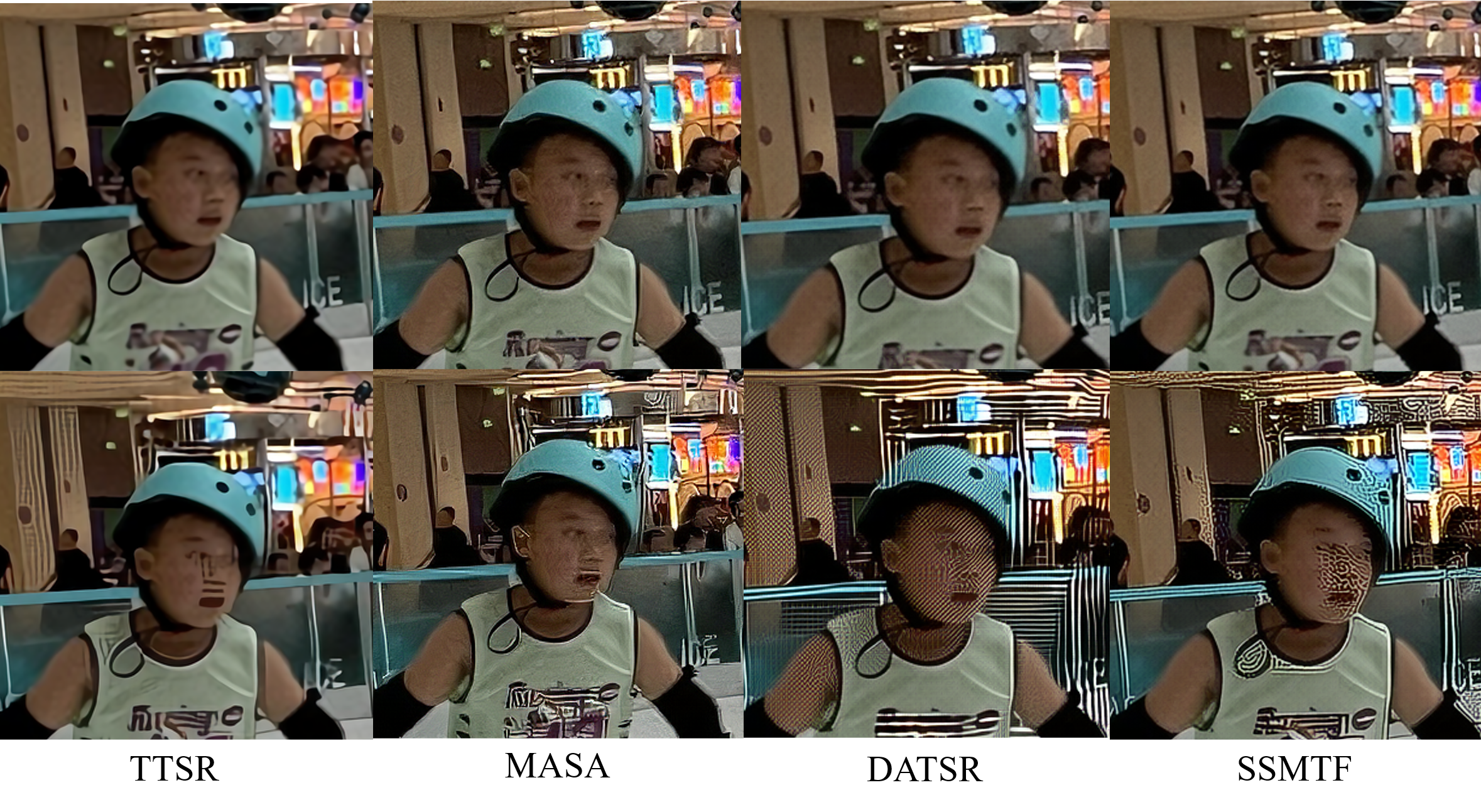}
\caption{Visual results on DRefSR dataset.}
\label{fig:vis_dref}
\end{figure}

To visually assess the impact of RefSR-Adv, we provide qualitative comparisons across the CUFED5, WR-SR, and DRefSR datasets in Figs. \ref{fig:vis_cufed}, \ref{fig:vis_wr}, \ref{fig:vis_dref}. 
For each victim model, we present a vertically aligned pair of super-resolved results: the top image represents the output generated using the original clean reference, while the bottom image illustrates the output synthesized under the perturbed adversarial reference.
Visual results demonstrate that this attack precisely disrupts the texture synthesis mechanism during super-resolution processing. While the global geometry of the generated image remains constrained by the low-resolution input, preventing complete collapse, high-frequency texture details guided by the reference image are severely compromised. Consequently, RefSR-Adv successfully induces significant texture illusions within the output, where previously coherent and valid semantic textures are systematically replaced by chaotic and perceptible visual artifacts. This specific disruption is notably more pronounced in advanced models designed for extreme detail restoration and high-fidelity texture migration, such as the DATSR and SSMTF. Furthermore, the consistency of these distortions across different data distributions further confirms the effectiveness and universality of the attack.
\section{Ablation Study}
In this section, we conduct a comprehensive ablation analysis to evaluate the key factors influencing the performance of RefSR-Adv. All experiments are performed on the CUFED5 using the full-loss version of the victim models.
\subsection{Impact of Perturbation Budget $\epsilon$}
As shown in Table \ref{tab:epsilon}, attack potency increases monotonically with perturbation budget $\epsilon$. However, $\epsilon=8/255$ provides the optimal balance between attacking performance and stealthiness (PSNR $>35$dB).
\begin{table}[h]
\centering
\caption{\footnotesize Ablation study of perturbation budget ($\epsilon$) with fixed iterations $T=50$. The chosen budget $\epsilon=8/255$ is highlighted in bold.}
\label{tab:epsilon}
\renewcommand{\arraystretch}{0.8}
\resizebox{\linewidth}{!}{%
\begin{tabular}{c|c|c|c|c}
\toprule
\textbf{Model} & \textbf{Budget ($\epsilon$)} & \textbf{Adversarial Output} & \textbf{Performance Drop} & \textbf{Stealthiness} \\
\midrule
\multirow{4}{*}{TTSR} 
& 2/255 & 24.36 / 0.7144 & 1.04 / 0.0456 & 46.52 / 0.9913 \\
& 4/255 & 23.02 / 0.6443 & 2.38 / 0.1157 & 40.93 / 0.9703 \\
& \textbf{8/255} & \textbf{21.84 / 0.5599} & \textbf{3.56 / 0.2001} & \textbf{35.71 / 0.9138} \\
& 16/255 & 20.85 / 0.4848 & 4.55 / 0.2752 & 30.99 / 0.8015 \\
\midrule
\multirow{4}{*}{MASA} 
& 2/255 & 23.35 / 0.7144 & 1.30 / 0.0113 & 48.05 / 0.9933 \\
& 4/255 & 21.58 / 0.6353 & 3.07 / 0.0904 & 42.38 / 0.9773 \\
& \textbf{8/255} & \textbf{19.69 / 0.5811} & \textbf{4.96 / 0.1446} & \textbf{36.99 / 0.9316} \\
& 16/255 & 18.86 / 0.5545 & 5.79 / 0.1712 & 32.13 / 0.8321 \\
\midrule
\multirow{4}{*}{DATSR} 
& 2/255 & 23.77 / 0.7361 & 3.99 / 0.0924 & 47.12 / 0.9925 \\
& 4/255 & 19.98 / 0.6010 & 7.78 / 0.2275 & 41.49 / 0.9745 \\
& \textbf{8/255} & \textbf{17.12 / 0.4690} & \textbf{10.64 / 0.3595} & \textbf{36.15 / 0.9236} \\
& 16/255 & 15.85 / 0.4030 & 11.91 / 0.4255 & 31.49 / 0.8204 \\
\midrule
\multirow{4}{*}{SSMTF} 
& 2/255 & 24.87 / 0.7607 & 3.26 / 0.0776 & 46.64 / 0.9918 \\
& 4/255 & 21.72 / 0.6607 & 6.41 / 0.1776 & 41.11 / 0.9731 \\
& \textbf{8/255} & \textbf{18.88 / 0.5569} & \textbf{9.25 / 0.2814} & \textbf{35.75 / 0.9193} \\
& 16/255 & 17.29 / 0.4948 & 10.84 / 0.3435 & 30.95 / 0.8069 \\
\bottomrule
\end{tabular}%
}
\end{table}
\subsection{Impact of Iteration Count $T$}
Table \ref{tab:iterations} indicates that while increasing iteration count $T$ slightly enhances the attack, $T=50$ is sufficient to achieve significant attacking performance with reasonable computational cost.
\begin{table}[h]
\centering
\caption{\footnotesize Ablation study of iteration steps ($T$) with fixed budget $\epsilon=8/255$. The chosen step $T=50$ is highlighted in bold.}
\label{tab:iterations}
\renewcommand{\arraystretch}{0.8}
\resizebox{\linewidth}{!}{%
\begin{tabular}{c|c|c|c|c}
\toprule
\textbf{Model} & \textbf{Iterations ($T$)} & \textbf{Adversarial Output} & \textbf{Performance Drop} & \textbf{Stealthiness} \\
\midrule
\multirow{4}{*}{TTSR} 
& 10 & 23.63 / 0.6601 & 1.77 / 0.0999 & 37.45 / 0.9374 \\
& 30 & 22.24 / 0.5858 & 3.16 / 0.1742 & 36.06 / 0.9200 \\
& \textbf{50} & \textbf{21.84 / 0.5599} & \textbf{3.56 / 0.2001} & \textbf{35.71 / 0.9138} \\
& 100 & 21.33 / 0.5286 & 4.07 / 0.2314 & 35.46 / 0.9088 \\
\midrule
\multirow{4}{*}{MASA} 
& 10 & 22.57 / 0.6552 & 2.08 / 0.0705 & 38.62 / 0.9491 \\
& 30 & 20.47 / 0.6029 & 4.18 / 0.1228 & 37.42 / 0.9371 \\
& \textbf{50} & \textbf{19.69 / 0.5811} & \textbf{4.96 / 0.1446} & \textbf{36.99 / 0.9316} \\
& 100 & 18.76 / 0.5545 & 5.89 / 0.1712 & 36.62 / 0.9261 \\
\midrule
\multirow{4}{*}{DATSR} 
& 10 & 22.15 / 0.6559 & 5.61 / 0.1726 & 37.98 / 0.9455 \\
& 30 & 18.45 / 0.5203 & 9.31 / 0.3082 & 36.59 / 0.9299 \\
& \textbf{50} & \textbf{17.12 / 0.4690} & \textbf{10.64 / 0.3595} & \textbf{36.15 / 0.9236} \\
& 100 & 15.76 / 0.4124 & 12.00 / 0.4161 & 35.80 / 0.9176 \\
\midrule
\multirow{4}{*}{SSMTF} 
& 10 & 22.52 / 0.6693 & 5.61 / 0.1690 & 37.51 / 0.9414 \\
& 30 & 19.82 / 0.5859 & 8.31 / 0.2524 & 36.10 / 0.9251 \\
& \textbf{50} & \textbf{18.88 / 0.5569} & \textbf{9.25 / 0.2814} & \textbf{35.75 / 0.9193} \\
& 100 & 18.16 / 0.5366 & 9.97 / 0.3017 & 35.54 / 0.9150 \\
\bottomrule
\end{tabular}%
}
\end{table}
\subsection{Impact of Reference Similarity}
To evaluate how the similarity between the low-resolution ($I_{LR}$) and reference ($I_{Ref}$) images affects attack performance, we conducted a comprehensive ablation study leveraging the five distinct similarity levels defined within the CUFED5 dataset. The quantitative results, as presented in Table \ref{tab:similarity}, demonstrate that at higher similarity levels (e.g., Level 1), RefSR models engage in more aggressive texture migration and feature fusion to maximize detail recovery. While this behavior is beneficial under benign conditions, it inadvertently facilitates the transmission and amplification of adversarial perturbations, leading to the most severe performance degradation. Conversely, at lower similarity levels (e.g., Level 5), the models' intrinsic correlation filtering mechanisms are more frequently triggered to reject mismatched features, which serves as a spontaneous and unintended defense that suppresses the propagation of adversarial noise. These observations indicate that the effectiveness of RefSR-Adv exhibits a significant positive correlation with the consistency between the $I_{LR}$ and $I_{Ref}$ input pairs. 
\begin{table}[h]
\centering
\caption{\footnotesize Ablation study of reference similarity levels (1 to 5) on CUFED5. Level 1 represents the highest similarity.}
\label{tab:similarity}
\renewcommand{\arraystretch}{0.8}
\resizebox{\linewidth}{!}{%
\begin{tabular}{c|c|c|c|c}
\toprule
\textbf{Model} & \textbf{Level} & \textbf{Adversarial Output} & \textbf{Performance Drop} & \textbf{Stealthiness} \\
\midrule
\multirow{5}{*}{TTSR} 
& \textbf{1} & \textbf{21.84 / 0.5599} & \textbf{3.56 / 0.2001} & \textbf{35.71 / 0.9138} \\
& 2 & 21.88 / 0.5584 & 3.42 / 0.1949 & 35.67 / 0.9152 \\
& 3 & 21.91 / 0.5545 & 3.27 / 0.1963 & 35.73 / 0.9120 \\
& 4 & 22.05 / 0.5592 & 3.12 / 0.1915 & 35.72 / 0.9134 \\
& 5 & 22.13 / 0.5576 & 3.11 / 0.1936 & 35.75 / 0.9086 \\
\midrule
\multirow{5}{*}{MASA} 
& \textbf{1} & \textbf{19.69 / 0.5811} & \textbf{4.96 / 0.1446} & \textbf{36.99 / 0.9316} \\
& 2 & 19.90 / 0.5940 & 4.52 / 0.1198 & 37.03 / 0.9327 \\
& 3 & 19.84 / 0.5909 & 4.53 / 0.1204 & 37.02 / 0.9298 \\
& 4 & 20.10 / 0.6000 & 4.22 / 0.1086 & 37.03 / 0.9310 \\
& 5 & 20.24 / 0.6089 & 4.10 / 0.0990 & 37.10 / 0.9280 \\
\midrule
\multirow{5}{*}{DATSR} 
& \textbf{1} & \textbf{17.12 / 0.4690} & \textbf{10.64 / 0.3595} & \textbf{36.15 / 0.9236} \\
& 2 & 17.31 / 0.5075 & 9.49 / 0.2878 & 36.34 / 0.9269 \\
& 3 & 17.42 / 0.5202 & 9.16 / 0.2659 & 36.41 / 0.9248 \\
& 4 & 17.54 / 0.5377 & 8.81 / 0.2398 & 36.45 / 0.9260 \\
& 5 & 17.56 / 0.5480 & 8.63 / 0.2204 & 36.51 / 0.9231 \\
\midrule
\multirow{5}{*}{SSMTF} 
& \textbf{1} & \textbf{18.88 / 0.5569} & \textbf{9.25 / 0.2814} & \textbf{35.75 / 0.9193} \\
& 2 & 19.43 / 0.6016 & 7.72 / 0.2044 & 35.94 / 0.9221 \\
& 3 & 19.60 / 0.6153 & 7.30 / 0.1817 & 36.01 / 0.9199 \\
& 4 & 19.88 / 0.6276 & 6.76 / 0.1588 & 36.05 / 0.9210 \\
& 5 & 20.10 / 0.6458 & 6.32 / 0.1310 & 36.13 / 0.9180 \\
\bottomrule
\end{tabular}%
}
\end{table}
\subsection{Comparison with Random Noise}
To confirm that the performance degradation is caused by specific adversarial perturbation rather than random noise, we compare RefSR-Adv with Gaussian noise at $\epsilon = 8/255$. Table \ref{tab:noise} shows that popular models are inherently robust to random noise. This confirms that RefSR-Adv can accurately exploit the model's dependence on reference features to induce severe artifacts, thus effectively distinguishing our targeted attacks from simple random noise interference.
\begin{table}[h]
\centering
\caption{\footnotesize Comparison with random noise on CUFED5. }
\label{tab:noise}
\renewcommand{\arraystretch}{0.8}
\resizebox{\linewidth}{!}{%
\begin{tabular}{c|c|c|c}
\toprule
\textbf{Model} & \textbf{Clean Output} & \textbf{Random Noise Output} & \textbf{Performance Drop} \\
\midrule
TTSR & 25.40 / 0.7600 & 25.39 / 0.7524 & 0.01 / 0.0076 \\
MASA & 24.65 / 0.7257 & 24.47 / 0.7092 & 0.18 / 0.0165 \\
DATSR & 27.76 / 0.8285 & 27.62 / 0.8181 & 0.14 / 0.0104 \\
SSMTF & 28.13 / 0.8383 & 27.93 / 0.8264 & 0.20 / 0.0119 \\
\bottomrule
\end{tabular}%
}
\end{table}
\section{POTENTIAL DEFENSE STRATEGIES}
To mitigate the identified threats, we suggest employing non-differential input purification, such as JPEG re-compression or bit-depth quantization to disrupt the high-frequency structures of adversarial perturbations, rendering them ineffective during the feature matching stage. Alternatively, a content-based matching gating mechanism could be introduced to block feature fusion when abnormal matching scores or semantic inconsistencies are detected. Furthermore, drawing on the findings by Huang \textit{et al.} \cite{huang2025scale}, adversarial fine-tuning can be utilized to force the model to learn more robust feature matching representations. 
\section{CONCLUSION}
This study reveals the security vulnerabilities of reference-based adversarial attacks in RefSR and proposes RefSR-Adv, a white-box attack framework targeting the reference image. Our results show that popular RefSR models are highly vulnerable to minute perturbations, which induce severe artifacts and degrade output quality. Crucially, we found a positive correlation between the similarity of the reference image and the attack success rate: higher-quality reference images exacerbate the model's vulnerability, confirming that over-reliance on reference features is a critical security flaw.

Despite its superior performance in white-box settings, the cross-model transferability of the attack remains challenging due to the architectural heterogeneity in feature matching and fusion mechanisms. Future work will focus on exploring black-box attacks by integrating meta-learning or query-based optimization, as well as developing similarity-aware defense mechanisms to enhance the robustness of RefSR systems.
\bibliographystyle{IEEEbib}
\bibliography{icme2026references}
\end{document}